\documentclass[10pt,twocolumn,letterpaper]{article}

\usepackage{cvpr}

\usepackage{epsfig}
\usepackage{graphicx}
\usepackage{amsmath}
\usepackage{amssymb}
\usepackage{url}
\usepackage{multirow,eucal}

\newcommand{\nframes}{{K}}
\newcommand{\ysubx}{{y_\mathbf{x}}}

\DeclareMathOperator*{\argmin}{arg\,min}

\usepackage[margin=4pt,font=footnotesize,labelfont=bf,labelsep=endash,tableposition=top]{caption}

\usepackage{cite}

\cvprfinalcopy %

\def\cvprPaperID{9712} %

\ifcvprfinal\fi
\begin{document}

\title{Self-trained Deep Ordinal Regression for End-to-End Video Anomaly Detection}

\author{
Guansong Pang$^{1*}$,
~~~
Cheng Yan$^{2,1}$\thanks{GP and CY equally contributed to this work. CY's contribution was made when visiting The University of Adelaide.},
~~~
Chunhua Shen$^1$\thanks{Corresponding author, e-mail: $ \sf chunhua.shen@adelaide.edu.au $},
~~~
Anton van den Hengel$^1$,
~~~
Xiao Bai$^2$
\\[0.125cm]
$ ^1$ The University of Adelaide, Australia
~ ~ ~ ~
$ ^2$
Beihang University, China
}

\maketitle
\thispagestyle{empty}

\begin{abstract}

Video anomaly detection is of critical practical importance to a variety of
real applications
because it allows human attention to be focused on events that are likely to be of interest, in spite of an otherwise overwhelming volume of video. We show that applying self-trained deep ordinal regression to video anomaly detection overcomes two key limitations of existing methods,
namely, 1)
being
highly
dependent
on manually labeled normal training data; and 2) sub-optimal feature learning. By formulating a surrogate two-class ordinal regression task we devise an end-to-end
trainable
video anomaly detection approach that enables joint representation learning and anomaly scoring without manually labeled normal/abnormal data. Experiments on eight real-world video scenes show that
our
proposed method outperforms state-of-the-art methods that require no labeled training data by a substantial margin, and enables easy and accurate localization of the identified anomalies.
Furthermore, we demonstrate
that our method offers effective human-in-the-loop anomaly detection which %
can be
critical in applications where anomalies are rare and the false-negative cost is high.

\end{abstract}

\section{Introduction}

\begin{figure}[h!]
  \centering
    \includegraphics[width=0.5\textwidth]{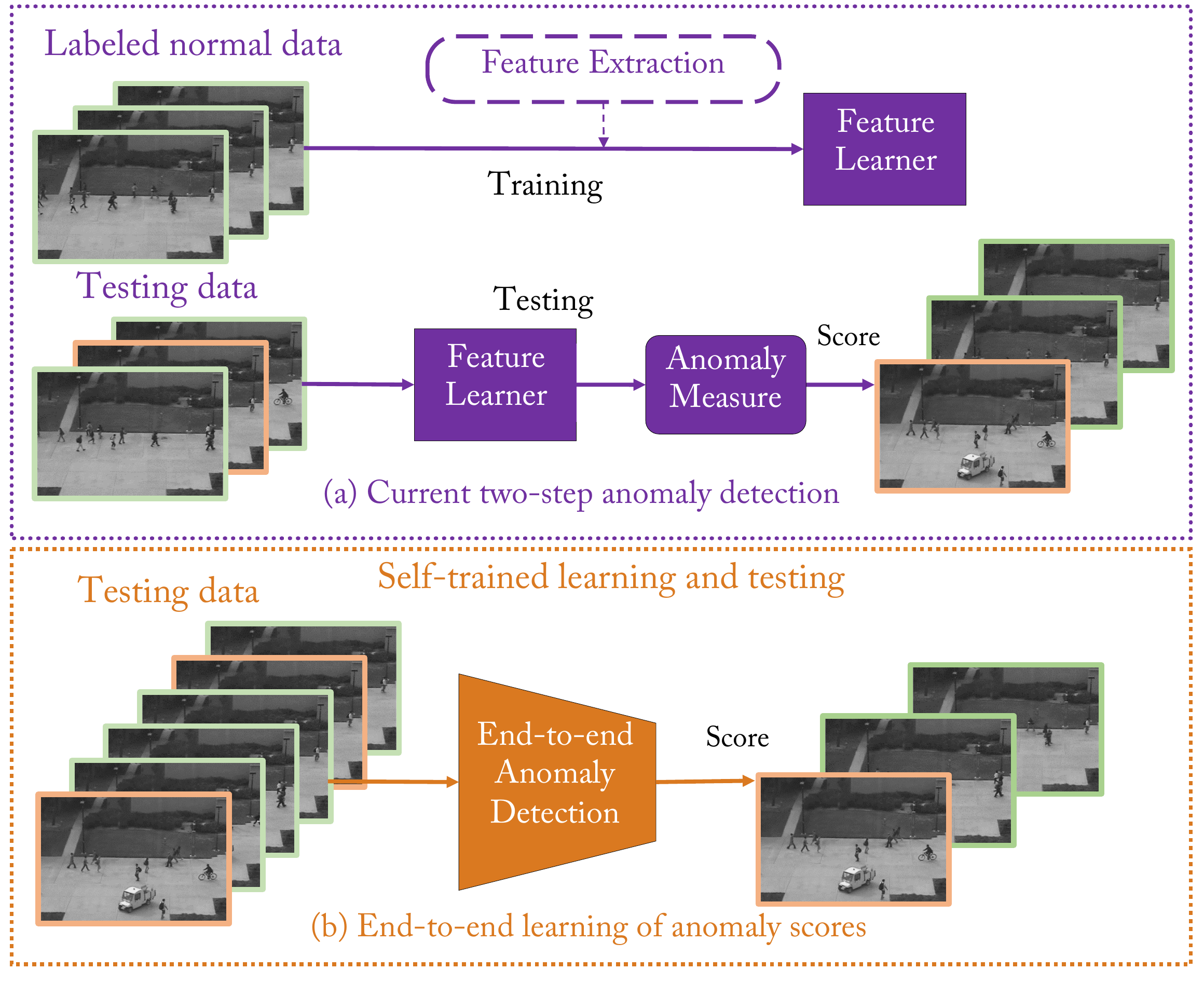}
  \caption{Pipelines of (a) two-step and (b) end-to-end anomaly detection. The two-step approach embodies separate feature extraction/learning and anomaly scoring methods, while the end-to-end approach unifies these two modules and directly learns the anomaly scores from the raw inputs. Also, the former often requires a set of labeled normal videos, whereas our approach requires no manual annotation of normal/abnormal data.}
  \label{fig:diff}
\end{figure}

Anomaly detection in video is the task of identifying frames from a video sequence which depict events that deviate significantly from the norm.
Identifying such anomalous events, \eg, fires, traffic accidents or stampedes,
can be of great practical importance, particularly in guiding timely responses.
The task is made particularly challenging, by the fact that anomalous events are rare, visually unbounded and
often unidentifiable before they occur.  It is nonetheless critical in a wide variety of application areas, as it provides a method for allowing human attention to be focused on the events that are most likely to be of interest from within an otherwise overwhelming volume of video.

The practical significance of the problem has seen many methods for video anomaly detection developed. The majority of existing methods assume the availability of a labeled dataset depicting a set of `normal' events.  This includes the dictionary learning-based methods \cite{cheng2015sparse,cong2011sparse,lu2013abnormal,zhao2011sparse} and the more recently emerging deep learning-based methods \cite{hasan2016ae,hinami2017fastrcnn,liu2018future,luo2017sparsernn,tudor2017unmasking,xu2015ae}. This assumption significantly limits their domain of application, not least because
it means
that
the system cannot be continuously re-trained without human involvement, nor applied to database sifting \cite{del2016discriminative}. To address this issue,
here we address
what has been termed `unsupervised video anomaly detection' \cite{del2016discriminative,tudor2017unmasking}, which requires identifying abnormal frames from a large volume of video frames with no manually labeled normal/abnormal training data. This version of the problem may be the only viable approach in the many application domains where human labeling of video is
extremely costly or
impossible.
This includes, for example,  large-scale video surveillance, Internet video filtering, and industrial process safety monitoring, because in all of these cases the definition of normal events is diverse, constantly changing, and unpredictable.

Additionally, these methods often employ a two-step approach as shown in Figure \ref{fig:diff}(a). This approach first learns or extracts feature representations from labeled normal training data and then employs a deterministic anomaly measure to calculate the normal/anomaly scores based on the extracted/learned representations. This essentially separates the feature learning/extraction and anomaly detection modules, leading to inflexible and sub-optimal anomaly scoring.
Some previous studies \cite{del2016discriminative,tudor2017unmasking} address a similar problem setting to ours. Although labeled normal training data is not required, their methods
typically
adopt a similar two-step approach that first extracts features from \textit{test data} and builds anomaly scoring models upon the extracted features.

To address the aforementioned two issues, we introduce an end-to-end approach to unsupervised video anomaly detection based on \textit{self-trained ordinal regression}.
Applying self-trained  ordinal regression enables the development of a weakly supervised approach to the problem that is amenable to end-to-end training.
End-to-end training, in turn, enables the optimization of a feature learner that is tailored specifically to anomaly detection in the target data, and an anomaly scoring process that is tailored to the output of this feature learner.
The method is weakly supervised in that it is initialized using a pre-trained model (\eg, ResNet-50 \cite{he2016deep}) on relevant auxiliary labeled data and initial pseudo labels of normality and abnormality generated using generic (that is, not video specific) anomaly detectors.

To apply self-training ordinal regression we formulate the problem as a surrogate two-class ordinal regression task, for which an ordinal regression \cite{gutierrez2016ordinal} model is defined to directly learn anomaly scores for pseudo normal and anomalous frames in an iterative fashion, as shown in Figure \ref{fig:diff}(b). The initial pseudo normal and anomalous frames can be determined based on existing unsupervised methods \cite{del2016discriminative,liu2012iforest,sugiyama2013sp,tudor2017unmasking}. The ordinal regression model stacks differentiable feature representations and anomaly scoring learners for end-to-end training. The key intuition underlying this approach is as follows. Although existing methods cannot produce well optimized anomaly scores, they generally achieve good accuracy in correctly identifying a subset of normal and anomalous events. These identified normal and anomalous events can be leveraged by the end-to-end anomaly score learner to iteratively improve and optimize the anomaly scores, resulting in significantly better detection performance compared to the initial detection.

We implement our formulation as a self-trained deep ordinal regression network, which is a synthesis of convolutional network-based feature learning and fully connected network-based anomaly scoring. This facilitates end-to-end anomaly score optimization. Moreover, our model can offer two important capabilities that are difficult to achieve with existing methods. First, the nature of self-training and end-to-end learning enables effective human-in-the-loop anomaly detection. Human experts can easily interact with our model by providing it with feedback on the detected anomalies which  the model can use to update itself to quickly return more accurate detection results. This is a critical capability, especially when the anomalies are rare and the cost of false-negatives is high. Second, our method can easily generate frame-level saliency maps to effectively localize the identified anomalies. In summary,
we make
the following two major contributions.
\begin{itemize}

\itemsep -0.12cm

    \item We show that applying self-training ordinal regression to video anomaly detection enables a novel formulation of the problem that not only eliminates the need for manually labeled training data, it enables end-to-end training, thus improving detection accuracy.

    \item We
    present
    a novel end-to-end neural network-based anomaly detection method. The method offers three critical capabilities:  i) it generates optimal anomaly scores w.r.t.\
    the given ordinal regression loss; ii) it enables effective human-in-the-loop anomaly detection; and iii) it offers easy and accurate localization of the identified anomalies within the corresponding images. These capabilities are analyzed through
    extensive
    experiments on eight different scenes from
    three real-world video
    datasets.

\end{itemize}

\subsection{Related Work}
Some popular video anomaly detection approaches include low-level feature extraction  \cite{adam2008umn,li2014anomaly,lin2016clustering,mahadevan2010anomaly,mehran2009abnormal,wu2010craft}, dictionary learning \cite{cheng2015sparse,chu2018sparse,cong2011sparse,lu2013abnormal,zhao2011sparse} and deep learning \cite{abati2019latent,chong2017ae,hasan2016ae,hinami2017fastrcnn,liu2018future,luo2017sparsernn,sabokrou2018adversarially,tudor2017unmasking,xu2015ae}.  The low-level feature extraction approach focuses on extracting low-level appearance \cite{adam2008umn}, and/or dynamic features \cite{li2014anomaly,mahadevan2010anomaly,mehran2009abnormal,wu2010craft}, for profiling normal behaviors. The dictionary learning-based approach learns a dictionary of normal events and identifies the events that cannot be well represented by the dictionary. The dictionary learning may also be applied to low-level features, such as histogram of gradient (HoG) or histogram of optical flow (HoF) features \cite{cong2011sparse,zhao2011sparse}, and 3D gradient features \cite{lu2013abnormal}. There have been other approaches that aim to model normal events with compact representations, such as hashing-based methods \cite{zhang2016hash} and clustering \cite{lin2016clustering,sun2017online}. The deep learning-based approach also aims  to learn a model of normal events\cite{lecun2015deep}. Most deep learning-based methods use reconstruction error to measure divergence of the test data from a set of normal training videos \cite{abati2019latent,chong2017ae,hasan2016ae,hinami2017fastrcnn,luo2017sparsernn,sabokrou2018adversarially,xu2015ae}. Future frame prediction is an alternative deep-learning approach explored in \cite{liu2018future}. The aforementioned approaches \cite{cheng2015sparse,chong2017ae,cong2011sparse,hasan2016ae,hinami2017fastrcnn,li2014anomaly,lin2016clustering,liu2018future,lu2013abnormal,luo2017sparsernn,sabokrou2018adversarially,sun2017online,xu2015ae,zhang2016hash,zhao2011sparse} often require manually labeled normal video samples to train their models. Unlike these approaches, our approach does not require manual annotation of normal/abnormal data.

Some recent %
work
\cite{del2016discriminative,tudor2017unmasking} %
addressed  a similar problem setting to ours. Permutation tests are used in \cite{del2016discriminative} to define the anomaly scores by evaluating the discriminativeness of a given frame in different groups of frames. Unmasking \cite{koppel2007unmasking} is exploited in \cite{tudor2017unmasking,liu2018unmasking} to measure the anomalousness by the classification accuracy changes resulting from unmasking. These methods generally include two main steps: feature extraction/learning and anomaly scoring, which are performed separately. This simplifies the feature extraction/learning problem but leads to a feature transform that is not optimized for the problem, or the data. By contrast, our approach unifies these two steps to enable end-to-end optimization of the whole anomaly identification process.
Studies
have
addressed  the same problem setting in the machine learning community, such as current state-of-the-art methods Sp \cite{sugiyama2013sp,pang2015lesinn} and iForest \cite{liu2012iforest}. Note that they operate on
traditional
features
and their effectiveness on video data is rarely explored.

Additionally, there has been some work for end-to-end anomaly detection \cite{hanson2018supervised,sultani2018mil,pang2019devnet}, but they require labeled normal and/or abnormal data for training.

\section{Problem Formulation}

The problem that we aim to address is end-to-end learning of anomaly scores for a set of video frames with no manually labeled normal/abnormal data. Formally, given a set of $\nframes$ video frames $\mathcal{X}=\{ \mathbf{x}_{1}, \mathbf{x}_{2}, \cdots, \mathbf{x}_{\nframes}\}$ with no class label information, our goal is to learn an anomaly scoring function $\phi: \mathcal{X} \mapsto \mathbb{R}$ that directly assigns anomaly scores to the video frames such that $\phi(\mathbf{x}_{i}) > \phi(\mathbf{x}_{j})$ if $\mathbf{x}_{i}$ is an anomalous frame and $\mathbf{x}_{j}$ is a normal frame.

We formulate this problem as a self-training ordinal regression task. Specifically, let $\mathcal{C}=\{c_1, c_2\}$ be \textit{augmented} scalar ordinal class labels with $c_1 > c_2$, $\mathcal{A} \subset \mathcal{X}$ be a set of anomalous frame candidates with each frame having an ordinal label $c_1$, $\mathcal{N} \subset \mathcal{X}$ ($\mathcal{N}\cap\mathcal{A} =\emptyset $) be a set of normal frame candidates with each frame having an ordinal label $c_2$, then the anomaly score learner $\phi$ can be formulated as
\begin{equation}\label{eqn:obj}
    \argmin_{\Theta} \sum_{\mathbf{x} \in
    \mathcal{G} } L\big(\phi(\mathbf{x};\Theta),
    \ysubx\big),
\end{equation}
where $\mathcal{G}=\mathcal{A} \cup \mathcal{N}$, $L(\cdot,\cdot)$ is a regression loss function and $\ysubx = c_1 \;,  \forall \mathbf{x} \in \mathcal{A}$ and $\ysubx = c_2 \;, \forall \mathbf{x} \in \mathcal{N}$.

One critical idea in the ordinal regression theory is to leverage the ordinal dependence in the supervision information to learn an optimal sample ranking function \cite{mccullagh1980ordinalregression,gutierrez2016ordinal}. To apply this idea to optimize the ranking of anomalies, we devise the self-training ordinal regression. As $c_1 > c_2$, optimizing the objective in Eqn. (\ref{eqn:obj}) will identify $\Theta^{*}$ corresponding to a version of $\phi(\mathbf{x};\Theta^{*})$ that assigns scores to $\mathcal{X}$ such that suspicious abnormal and normal samples have anomaly scores as close to  respective $c_1$ and $c_2$ as possible, yielding an optimal anomaly ranking.

Since $\mathcal{X}$ contains high-dimensional samples, we often need to map the data onto a low-dimensional space before the anomaly scoring. Let $\psi$ be a feature mapping function with parameters $\Theta_{r}$ and $\eta$ be an anomaly scoring function  with parameters $\Theta_{s}$, then Eqn. (\ref{eqn:obj}) can break down into

\begin{equation}\label{eqn:obj_psi}
    \argmin_{\Theta_{r},\Theta_{s}} \sum_{\mathbf{x} \in
    \mathcal{G}} L\big(\eta(\psi(\mathbf{x};\Theta_{r});\Theta_{s}), \ysubx\big).
\end{equation}

To enable end-to-end training, we need to address two main problems. First, it requires that $\phi$ and $\psi$ can be simultaneously optimized. As is discussed in Section \ref{subsec:end2end}, deep neural networks can be designed to address this problem. Second, we need to generate $\mathcal{A}$ and $\mathcal{N}$ since they are unknown in the first place. To tackle this problem, we first initialize $\mathcal{A}$ and $\mathcal{N}$ using anomaly scores generated by some existing unsupervised anomaly detection methods (see Section \ref{subsec:initial}), and we iteratively update $\mathcal{A}$ and $\mathcal{N}$ and retrain $\phi$ until the best $\phi$ is achieved (see Section \ref{subsec:selftraining}).

The key idea of this formulation is that existing anomaly detection methods are able to identify frames that are clearly members of $\mathcal{A}$ and $\mathcal{N}$, but are unable to expand these sets to accurately cover $\mathcal{X}$.
The proposed approach uses each estimate of $\mathcal{A}$ and $\mathcal{N}$ to further optimize the anomaly scores. These scores then, in turn, help generate new, more accurate, sets $\mathcal{A}$ and $\mathcal{N}$. This iterative self-training achieves much better detection performance and coverage than the initial detection methods support.

It is important to note that the initial set $\mathcal{A}$ need not span the anomalies in $\mathcal{X}$, and may even erroneously contain elements of $\mathcal{N}$. The iterative process means that the model implicitly defined by $\Theta$ is continually being refined. The membership of $\mathcal{A}$ and $\mathcal{N}$ is thus constantly being updated to improve their quality.

\section{The Proposed Method}

\begin{figure*}[h!]
  \centering
    \includegraphics[width=0.72\textwidth]{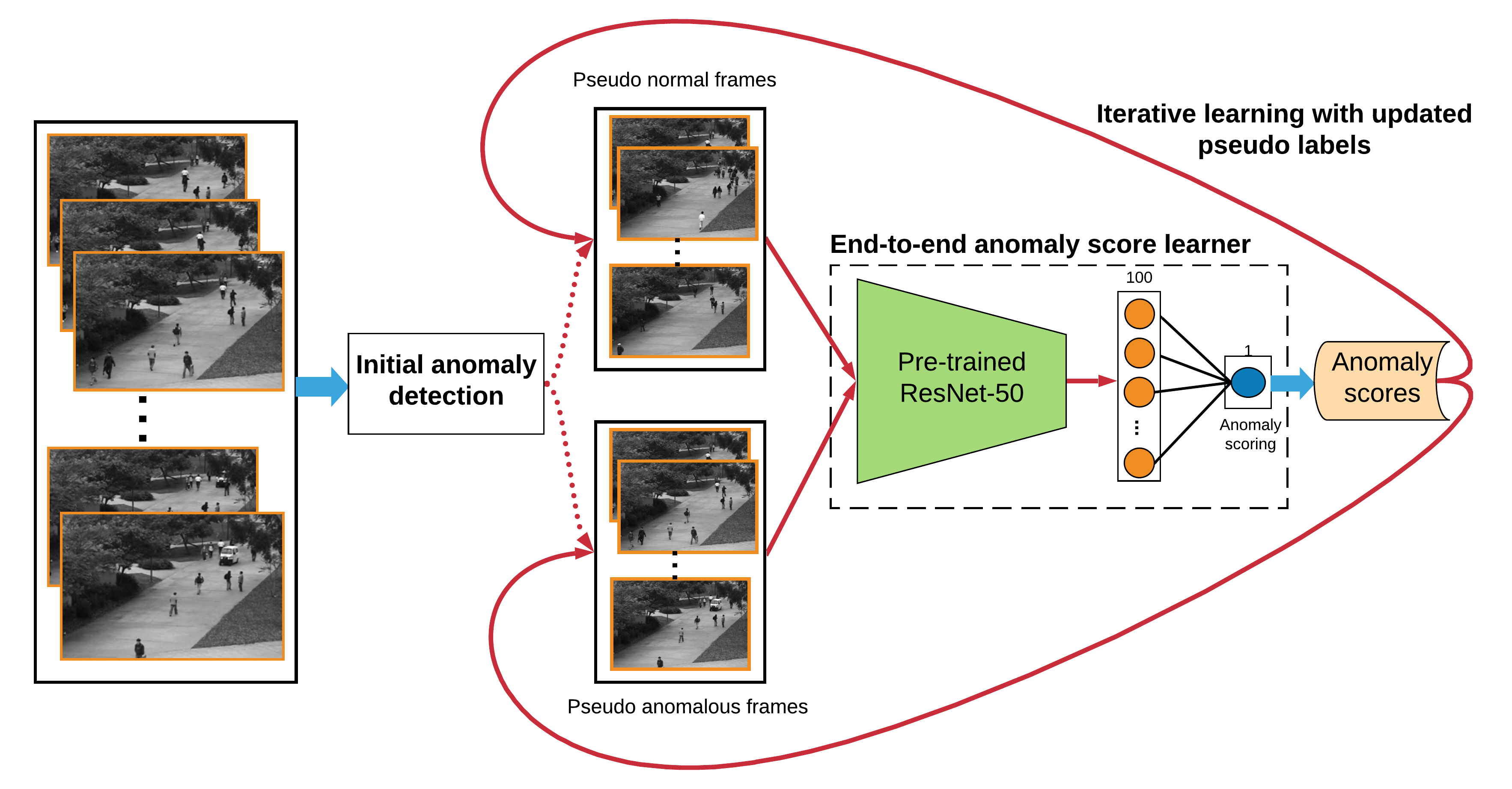}
  \caption{The proposed framework. Given a set of unlabeled videos, we first perform initial detection to generate pseudo anomalous and normal frame sets. These sets are then used to train a (pre-trained) ResNet-50 model \cite{he2016deep} and a fully connected network in a end-to-end fashion. The trained model is then used to recompute the anomaly scores of all frames. The membership of $\mathcal{A}$ and $\mathcal{N}$ is updated accordingly, and the process repeated.}
  \label{fig:framework}
\end{figure*}

Our formulation is implemented as a self-training deep neural network for ordinal regression.
As shown in Figure \ref{fig:framework}, our method consists of three major modules. The first module carries out the initial anomaly detection, which yields the initial membership of $\mathcal{A}$ and $\mathcal{N}$. These are then fed into the end-to-end anomaly scoring module to optimize $\Theta$ and thus the anomaly scores. A corresponding new set of anomaly scores are then generated, which are used to update the membership of $\mathcal{A}$ and $\mathcal{N}$.

\subsection{Initial Anomaly Detection}\label{subsec:initial}

The role of the initial anomaly detection is to obtain a set of frames that can be identified as belonging to $\mathcal{A}$ and $\mathcal{N}$ with high probability. To achieve this, two state-of-the-art unsupervised anomaly detection methods, Sp \cite{sugiyama2013sp,pang2015lesinn} and iForest \cite{liu2012iforest}, which are designed for feature vector-based data, are employed to perform the initialization. Combining anomaly scores from different anomaly detectors allows to identify anomalies with different characteristics and improves the subsequent anomaly candidate selection. Sp is a very simple yet effective and provable method, which defines the anomaly score of a given example as the nearest neighbor distance in a small random subsample of the full dataset. Let $f(\cdot)$ be a function that extracts feature vectors from a video frame and $\mathbf{z}=f(\mathbf{x})$, its anomaly score is
\begin{equation}\label{eqn:knnscore}
    s_{1}(\mathbf{z}) = \min_{\Tilde{\mathbf{z}}\in\mathcal{S}} d(\mathbf{z},\Tilde{\mathbf{z}}),
\end{equation}
where $\mathcal{S} \subset \mathcal{X} $ is a random subset of $\mathcal{X}$ and $d(\cdot,\cdot)$ represents the Euclidean distance.
To obtain statistically stable performance, for each $\mathbf{z}$, we use the average of a bootstrap aggregating of $m$ scores to be the final anomaly score \cite{pang2015lesinn}.

iForest \cite{liu2012iforest} is probably the most widely-used unsupervised anomaly detection method for traditional general multidimensional data in recent years. It posits that anomalies are susceptible to isolation and builds isolation trees on random data subspaces to identify them. Each tree is grown by using a random subsample until every data example is isolated, where the feature and cut-point at each tree node are randomly selected. The inverse of the path length traversed from the root to a leaf node by $\mathbf{x}$ is used as its anomaly score
\begin{equation}\label{eqn:iforest}
  s_{2}(\mathbf{z})=2^{\frac{E(h(\mathbf{z}))}{c(|\mathcal{S}|)}} ,
\end{equation}
\noindent where $h(\mathbf{z})$ denotes the path length of $\mathbf{z}$ in $\mathcal{S}$, $E(h(\mathbf{z}))$ is the average path length of $\mathbf{z}$ over $n$ isolation trees, and $c(\cdot)$ is the expected path length for a given subsample size.

We rescale the two sets of anomaly scores from these two detectors into the same range and use their average as the initial anomaly scores. We then use these anomaly scores to include the most likely anomalous frames into the pseudo anomaly set $\mathcal{A}$ and the most likely normal frames into the pseudo normal set $\mathcal{N}$ (see Section \ref{sec:imp} for detail).

\subsection{End-to-end Anomaly Score Learner}\label{subsec:end2end}
The end-to-end anomaly score learner takes $\mathcal{A}$ and $\mathcal{N}$ as inputs and learns to optimize the anomaly scores such that the data inputs of similar behaviors as those in $\mathcal{A}$ ($\mathcal{N}$) receive large (small) scores. The score learner can be defined as a function $\phi(\cdot; \Theta):\mathcal{X} \mapsto \mathbb{R}$, which is a sequential stack of a \textit{feature representation learner} $\psi(\cdot; \Theta_{r}): \mathcal{X} \mapsto \mathcal{Q}$ and an \textit{anomaly score learner} $\eta(\cdot; \Theta_{s}): \mathcal{Q} \mapsto \mathbb{R}$, where $\mathcal{Q} \in \mathbb{R}^{M}$ is an intermediate feature representation space and $\Theta=\{\Theta_{r}, \Theta_{s}\}$ contains all the parameters to be learned.

Specifically, the feature learner $\psi(\cdot; \Theta_{r})$ is specified as a network with $H \in \mathbb{N}$ hidden layers and their weights $\Theta_{r}=\{\mathbf{W}^{1}, \mathbf{W}^{2}, \cdots, \mathbf{W}^{H}\}$, and can be represented as
\begin{equation}
    \mathbf{q} = \psi(\mathbf{x}; \Theta_{r}),
\end{equation}
where $\mathbf{x} \in \mathcal{X}$ and $\mathbf{q} \in \mathcal{Q}$. Different backbones can be used here. We implement $\psi$ using ResNet-50 \cite{he2016deep} due to its superior capability of capturing frame appearance features.

The anomaly score learner $\eta(\cdot, \Theta_{s}): \mathcal{Q} \mapsto \mathbb{R}$ is specified as a fully connected two-layer neural network. The network consists of a hidden layer with 100 units and an output layer with a single linear unit:
\begin{equation}
    \eta(\mathbf{q};\Theta_{s}) = \mathbf{w}^\intercal g(\mathbf{q}; \mathbf{W}^{H+1}),
\end{equation}
where $\mathbf{q} \in \mathcal{Q}$, $[\cdot]^\intercal$ is a matrix transpose operation,  $g(\cdot)$ maps the ResNet-50 features to the hidden layer and $\Theta_{s} = \{\mathbf{W}^{H+1}, \mathbf{w}\}$ contains the weight parameters of this scoring learner. Thus, $\phi(\cdot; \Theta)$ can be formally represented as
\begin{equation}
    \phi(\mathbf{x};\Theta) = \eta\big(\psi(\mathbf{x};\Theta_{r});\Theta_s\big),
\end{equation}
which directly maps raw visual inputs to scalar anomaly scores and can be trained in an end-to-end fashion by minimizing the following loss function:

\begin{equation}\label{eqn:loss}
   L\big(\phi(\mathbf{x};\Theta), \ysubx\big) = |\phi(\mathbf{x};\Theta) - \ysubx|,
\end{equation}

\noindent where $\ysubx=c_1$ when $\mathbf{x} \in \mathcal{A}$ and $\ysubx=c_2$ when $\mathbf{x} \in \mathcal{N}$. Since $y$ takes two scalar ordinal values only, it is a two-class ordinal regression.

The absolute loss is employed together with stochastic gradient descent-based optimization to reduce the negative effects brought by false pseudo labels in $\mathcal{A}$ and $\mathcal{N}$. Minimizing this loss enforces an anomaly score close to $c_1$ ($c_2$) for any frames having similar features as the frames in $\mathcal{A}$ ($\mathcal{N}$), resulting in larger anomaly scores assigned to anomalous frames than normal frames.

\subsection{Iterative Learning via Self-training}\label{subsec:selftraining}

We further perform iterative learning using a self-training approach \cite{zhu2009selftraining} to iteratively improve our anomaly detector. The intuition is that the initial anomaly detection results can pose limitations on the performance of our end-to-end anomaly score learner, since our score learner is dependent on the quality of pseudo normal and anomalous frames; on the other hand, our end-to-end anomaly score optimization is expected to produce better anomaly scores than the initial anomaly scores, so it can provide pseudo normal and anomalous frames of better quality and in turn improve its self-performance.

Self-training, a.k.a.\  self-learning, is a classic semi-supervised learning approach. It first trains a model using a small labeled dataset and then applies the trained model to unlabeled data to generate more reliable labeled data. Since we do not have any labeled data, we present a simple strategy to adapt the self-training to an unsupervised setting. Particularly, at each iteration of the iterative learning, instead of incrementally adding more labeled data, we use the newly obtained pseudo labels, $\mathcal{A}$ and $\mathcal{N}$, to replace the previous ones and then retrain the end-to-end anomaly learner $\phi$. The main reason for discarding the previous $\mathcal{A}$ and $\mathcal{N}$ is because combining the previous and newly obtained pseudo labels without supervision information can result in worse pseudo labels. We found empirically that this simple strategy worked very well on different datasets.

Each iteration outputs an optimized $\phi$, so the iterative learning results in a set of trained models. Similar to sequential ensemble learning \cite{zhou2012ensemble}, we perform an average aggregation of all the sequentially output models to achieve stable detection performance. Specifically, the final anomaly score of a given $\mathbf{x}$ is defined as
\begin{equation}
    \text{score}(\mathbf{x}) = \frac{1}{t}\sum_{i=1}^{t}\phi_{i}(\mathbf{x}),
\end{equation}
where $\phi_{i}$ is the optimized model at the $i$th iteration.

\section{Experiments}

\subsection{Implementation Details}\label{sec:imp}
The initial anomaly detectors, Sp and iForest, are implemented using scikit-learn.
They are used with recommended settings as in \cite{liu2012iforest,pang2015lesinn}. Since both Sp and iForest only work on feature vectors, we need to transform video data into feature vectors before applying them. Specifically, we first extract features using the last dense layer of the pre-trained ResNet-50 %
and then apply PCA to reduce the dimensionality using the most important 100 components.

For the end-to-end anomaly score learner, the pre-trained ResNet-50\footnote{ResNet-50 is used since this work examines appearance-based anomalies only, which is one of the most common anomalies in video data.
} is used as our feature learner; the fully connected 100-unit hidden layer uses the ReLU activation function $a(u) = \mathit{max}(0, u)$; the output layer contains a linear unit; and $c_1=1$ and $c_2=0$ are used in the pseudo ordinal class labels to guide the learning (our model also worked well with other settings of $c_1$ and $c_2$ as long as $c_1$ was sufficiently larger than $c_2$). The Stochastic Gradient Descent (SGD) optimizer with a learning rate 0.005 is used  throughout all the experiments. The batch size and the number of epochs are respectively set to 128 and 50 by default.

To obtain a set of reliable pseudo anomalous frames, we need to determine $\mathcal{A}$ with a sufficiently high confidence level. Particularly, we include the 10\% most anomalous frames into $\mathcal{A}$ according to their anomaly scores, because anomaly scores often follow a Gaussian distribution \cite{kriegel2011interpreting} and this decision threshold can provide an approximate 90\% confidence level of making false positive errors in such cases. However, we may still include normal frames into $\mathcal{A}$. We further tackle this problem using a weighted random sampling-based mini-batch generation method, i.e., sampling examples from $\mathcal{A}$ with a probability positively proportional to their anomaly scores. To generate the pseudo normal frame set $\mathcal{N}$, we select the 20\% most normal frames based on the anomaly scores. This can always help to achieve a high-quality $\mathcal{N}$ because of the overwhelming presence of normal frames in the real-word datasets. These two cutoff thresholds are used by default as they consistently obtain substantially improved performance on datasets with diverse anomaly rates (see Table \ref{tab:auconthreedata} and Figure \ref{fig:initial} for detail).

In the iterative learning, extensive results showed that our model could often be substantially improved in the first multiple iterations and then plateaued out. Thus, we perform the iterative learning for five iterations by default.

\subsection{Datasets}

Three real-world datasets are used in our experiments:

\begin{itemize}

\itemsep -0.112cm

    \item \textbf{UCSD} \cite{mahadevan2010anomaly}. This data is one of the most challenging anomaly detection datasets. It contains the UCSD Pedestrian 1 data (Ped1) and the UCSD Pedestrian 2 data (Ped2). Ped1 consists of 34 training and 36 test videos, while Ped2 contains 16 training and 12 test videos. The anomalies are vehicles, bicycles, skate borders and wheelchairs crossing the pedestrian areas.

    \item \textbf{Subway} \cite{adam2008umn}. This is one of the largest datasets for video anomaly detection. It includes two videos: the Entrance gate video of 96 minutes and the Exit gate video of 43 minutes. The anomalies are passengers walking towards a wrong direction or escaping tickets.

    \item \textbf{UMN} \cite{umndata}. The UMN dataset contains three different scenes, each with 1,453, 4,144, and 2,144 frames, respectively. In each scene, the normal activity is people casually walking around while the anomalous activity is people running in all directions.
\end{itemize}

Note that anomalies are rare events in real-world applications, but this is violated if only the test set of these datasets are used, because these test sets can contain a large percentage of anomalous events, \eg, nearly 50\% anomalies in the UCSD test sets. Such test sets are not applicable in our setting. We address this problem by merging the training and test sets, and train and evaluate our model on the full dataset with the ground truth used in the evaluation only. This applies to the competing methods unless stated otherwise.

\subsection{Performance Evaluation Metrics}

Following previous work \cite{del2016discriminative,liu2018future,luo2017sparsernn,sugiyama2013sp,sultani2018mil,tudor2017unmasking}, the area under the ROC curve (AUC) is used as the evaluation metric. AUC is calculated using the frame-level anomaly scores and ground truth. The equal error rate (EER) has also been used as the evaluation metric in some previous work, but we agree with \cite{del2016discriminative} that this metric can be misleading for many real-world applications where the anomalies are very rare. Therefore, we do not use the EER in our evaluation.

\subsection{Effectiveness in Real-world Datasets}
As shown in Table \ref{tab:auconthreedata}, we first examine the performance of our method by comparing to 17 state-of-the-art methods. On all of the scenes, our method is consistently the best performer among the unsupervised methods \cite{liu2012iforest,sugiyama2013sp,del2016discriminative} that are evaluated using the same evaluation protocol. Specifically, compared to Sp + iForest \cite{liu2012iforest,sugiyama2013sp} that provides initial anomaly detection results for our method, our method achieves about 2\%-15\% AUC improvement. It is impressive that we obtain more than 15\% improvement on UCSD-Ped1 and UCSD-Ped2 where Sp + iForest works less effectively, and that we can also obtain remarkable 8\%-12\% improvement on different scenes of the UMN data where Sp + iForest performs very well. This demonstrates our method's capability in producing significantly better anomaly scores than the initial scores when the initial detectors perform either fairly well or very well. To have a fair and straightforward comparison with the discriminative framework-based method \cite{del2016discriminative}, our method is compared with its two variants: the first variant, namely Del Giorno et al. \cite{del2016discriminative} \#1 in Table \ref{tab:auconthreedata}, uses ResNet-50 and PCA to extract features as the input to the discriminative framework; and the second variant, namely Del Giorno et al. \cite{del2016discriminative} \#2 in Table \ref{tab:auconthreedata}, uses the features extracted from the last dense layer in our trained model. The results show that our method achieves about 5\%-25\% AUC improvement over both cases of Del Giorno et al. \cite{del2016discriminative} on all the datasets.

\begin{table*}[htbp]
\centering
\caption{Frame-level AUC performance. Our method is compared with 12 methods that require labeled normal data in the upper block and five methods that require no labeled normal/abnormal data in the bottom block. The best performance in each block is boldfaced.}
\scalebox{0.81}{
\begin{tabular}{ l| r |cc|cc|cccc }
\hline
&& \multicolumn{2}{|c | }{\textbf{UCSD}} & \multicolumn{2}{|c|}{\textbf{Subway}} & \multicolumn{4}{c }{\textbf{UMN}}
        \\
        \hline
\textbf{Training Data} &\textbf{Method} & \textbf{Ped1} & \textbf{Ped2} & \textbf{Entrance} & \textbf{Exit} & \textbf{Scene1} & \textbf{Scene2} & \textbf{Scene3} & \textbf{All Scenes} \\ \hline
\multirow{12}{9em
}{{Labeled normal data}} &Kim et al. \cite{kim2009observe} & 59.0\% & 69.3\% & - & - & - & - &- & -\\
&Mahadevan et al. \cite{mahadevan2010anomaly} & 81.8\% & 82.9\% & - & - & - & - & - &-\\
&Mehran et al. \cite{mehran2009abnormal}  & 67.5\% & 55.6\% & - & - & - & - & - & 96.0\% \\
&Cong et al. \cite{cong2011sparse} & - & - & 80.0\% & 83.0\% & 99.5\% & 97.5\% & 96.4\% &97.8\% \\
&Xu et al. \cite{xu2015ae} & 92.1\% & 90.8\% & - & - & - & - & - & - \\
&Sun et al. \cite{sun2017online}  & \textbf{93.8\%} & 94.1\% & - & - & 99.8\% & \textbf{99.3\%} & \textbf{99.9\%} & \textbf{99.7\%} \\
&Zhang et al. \cite{zhang2016hash}  & 87.0\% & 91.0\% &-  & - & 99.2\% & 98.3\% & 99.9\% & 99.7\% \\
&Liu et al. \cite{liu2018future} & 83.1\% & 95.4\% & -  & - & - & - & - & - \\
&Nguyen et al. \cite{nguyen2019anomaly} & - & 96.2\% & -  & - & - & - & - & - \\
&Dong et al. \cite{gong2019memorizing} & - & 94.1\% & -  & - & - & - & - & - \\
&Ionescu et al. \cite{ionescu2019object} & - & \textbf{97.8\%} & -  & - & - & - & - & 99.6\% \\
& Ionescu et al. \cite{ionescu2019wacv} & - & -&  \textbf{93.5\%} & \textbf{95.1\%} & \textbf{99.9\%} & 98.2\% & 99.8\% & 99.3\% \\
\hline
\multirow{6}{9em}{   {No labeled data}  } & Ionescu et al. \cite{tudor2017unmasking} & 68.4\% & 82.2\% & 70.6\% & 85.7\% & 99.3\% & 87.7\% & 98.2\% & 95.1\% \\
& Liu et al. \cite{liu2018unmasking} & 69.0\% & \textbf{87.5\%} & 71.6\% & \textbf{93.1\%} & - & -& -&95.2\%\\
&Del Giorno et al. \cite{del2016discriminative} \#1& 50.3\% & 63.0\% & 70.7\% & 86.8\% & 82.5\% & 83.5\% & 87.4\% & 76.5\% \\
&Del Giorno et al. \cite{del2016discriminative} \#2 & 59.6\% & 57.6\% & 74.6\% & 87.2\% & 80.2\% & 88.3\% & 77.1\% & 84.8\% \\
&Sp + iForest \cite{liu2012iforest,sugiyama2013sp} & 56.3\% & 67.5\% & 80.5\% & 91.0\% & 87.3\% & 88.1\% & 91.5\%  & 87.1\% \\
&\textbf{Our method} & \textbf{71.7\%} & 83.2\% & \textbf{88.1\%} & 92.7\% & \textbf{99.9\%} & \textbf{99.9\%} & \textbf{99.7\%}  & \textbf{97.4\%} \\ \hline
\end{tabular}
}
\label{tab:auconthreedata}
\end{table*}

Compared to the unmasking framework \cite{tudor2017unmasking}, our method performs substantially better on the challenging cases UCSD-Ped1, Subway-Entrance, Subway-Exit and UMN-Scene2 by a margin of respective 3\%, 17\%, 7\% and 12\%, and performs comparably better on the other datasets. The unmasking method is improved by a two-sample test method in \cite{liu2018unmasking}. Our method retains similar improvement over this variant on all scenes except UCSD-Ped2 and Subway-Exit. Note the results of the unmasking method and its variant are respectively taken from \cite{tudor2017unmasking} and \cite{liu2018unmasking} based on an evaluation protocol different from ours, i.e., they are evaluated on the test data rather than the full data.

Compared to the upper block methods, it is very impressive that our method
1) achieves large improvement over several of these methods, such as Kim et al.~\cite{kim2009observe}, Mahadevan et al. \cite{mahadevan2010anomaly} and Mehran et al. \cite{mehran2009abnormal} on the UCSD data and Cong et al. \cite{cong2011sparse} on the Subway and UMN datasets;
and 2) performs comparably well to the best methods \cite{sun2017online,zhang2016hash,ionescu2019object,ionescu2019wacv} on the UMN data. However, it is clear that our AUC score is 10\%-22\% lower than the methods \cite{liu2018future,sun2017online,zhang2016hash} on the UCSD data, showing the substantial gap between these two types of methods on the very challenging data. Note that the upper block competing methods are also based on a different setting from ours and taken here for a high-level comparison only.

\subsection{Human-in-the-loop Anomaly Detection}
Existing methods are lack of explicit prior knowledge of anomalies. As a result, many anomalous events they identify are data noises. This section examines whether our method can effectively interact with human experts to leverage their feedback on the anomalies of their interest to iteratively enhance our model and reduce such false positives. We simulate the interaction as follows. First, our model presents a small set of $l$ top-ranked anomalous frames to an expert. The expert then picks up two sets of $k$ frames ($k \ll l$), with one set of $k$ frames believed to be the anomalies of interest and another set to be normal events. These frames are employed to fine-tune our model with 20 epochs. After that, an updated anomaly ranking is again presented to the expert for the feedback. This human-machine interaction can repeat until the expert obtains the most satisfactory anomaly ranking results. To better exploit the feedback, we also include the temporally adjacent frames of the selected frames with the same label (\eg, the adjacent 5 frames to the selected frame) into our fine-tuning process.

\begin{figure*}[h!]
  \centering
    \includegraphics[width=0.7648\textwidth]{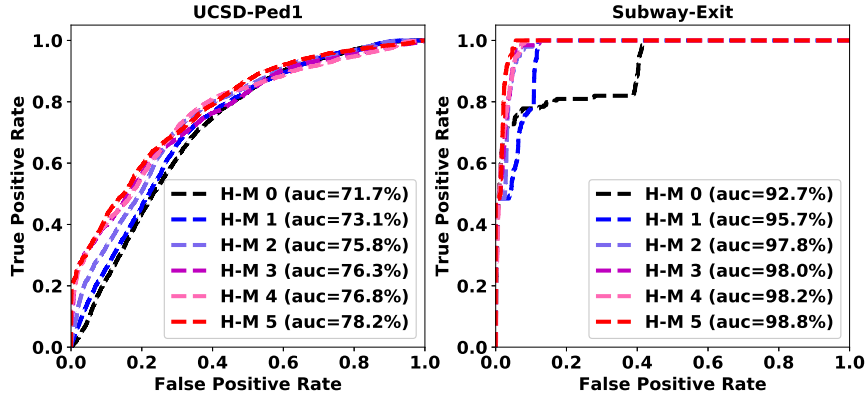}
  \caption{ROC curves of our human-in-the-loop anomaly detection. H-M $i$ indicates the results obtained in the $i$-th human-machine interaction. Best viewed in color.}
  \label{fig:feedback}
\end{figure*}

Figure \ref{fig:feedback} shows empirical results on two representative datasets, UCSD-Ped1 and Subway-Exit, with $l=\lfloor0.1N\rfloor$ and $k=5$. UCSD-Ped1 represents a challenging scene with large improvement space, while Subway-Exit represents a less challenging scene yet with small improvement space. Our method can well leverage the limited human feedback per interaction to gradually and consistently reduce the false positive errors, achieving more than 6\% AUC improvement on both datasets after 5-round interactions.

\subsection{Localizing the Identified Anomalies}
\begin{figure*}[t!]
  \centering
    \includegraphics[width=0.842\textwidth]{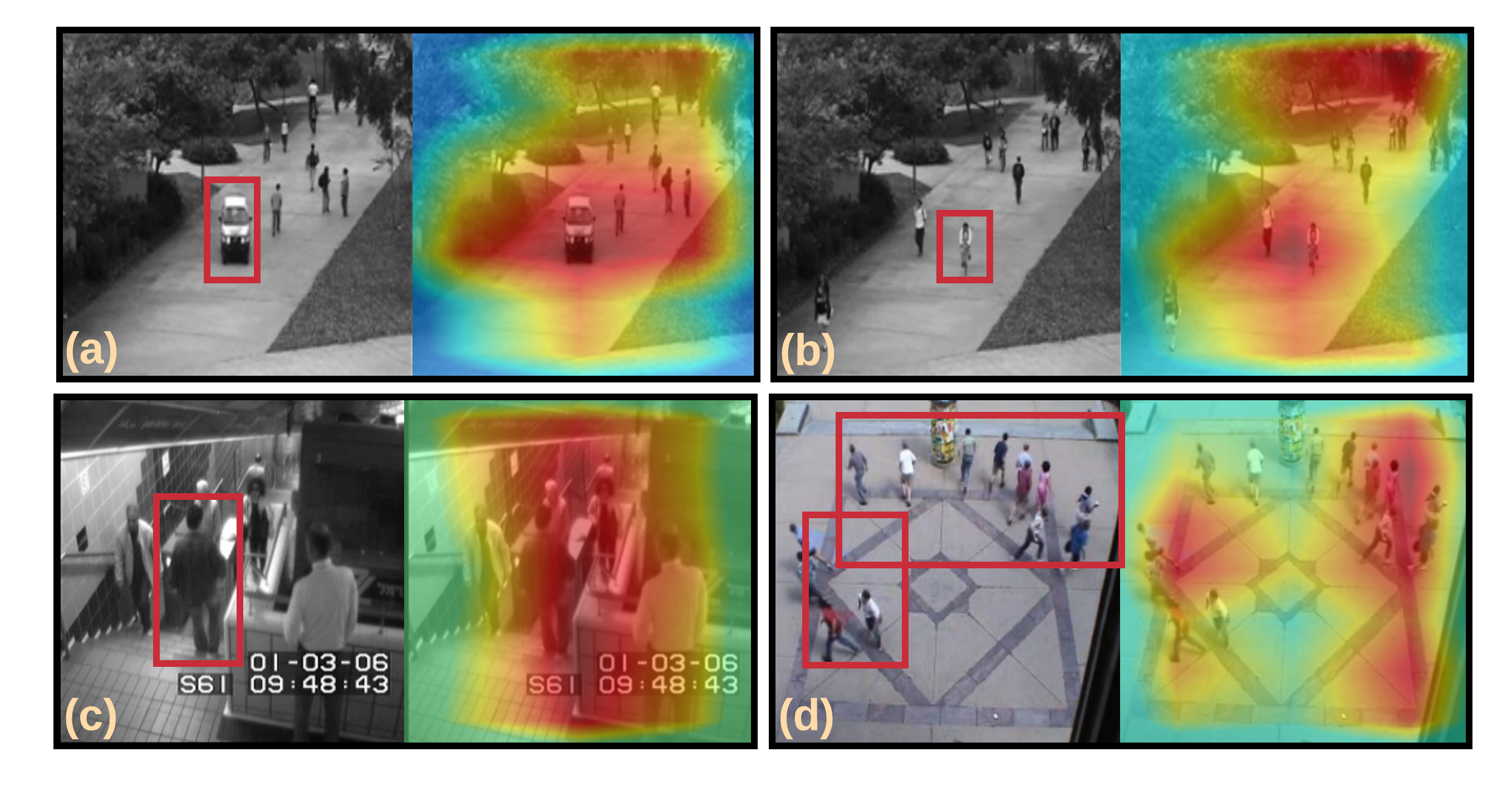}
  \caption{Manually labeled exemplar anomalies in the original inputs (red rectangles on the left) and the corresponding CAM \cite{zhou2016cam} based saliency maps yielded by our method (right). Anomalies in subfigures (a) and (b) are respectively vehicles and bikes crossing the pedestrian areas in the UCSD data. The anomaly in subfigure (c) is from the Subway-Exit data, a passenger walking towards a wrong direction. Subfigure (d) shows the anomaly from the UMN data, people running in all directions.   }
  \label{fig:saliency}
\end{figure*}

The end-to-end learning of anomaly scores also enables us to leverage existing deep neural network interpretation techniques to localize and understand the anomalous patches within a given frame that are responsible for large anomaly scores. Here we adapt the state-of-the-art method, class activation map (CAM) \cite{zhou2016cam}, to achieve this goal. Particularly, for a given frame $\mathbf{x}$, let $p_{k}(i,j)$ be the activation of an unit $k$ in the last convolutional layer at a spatial location $(i,j)$ and $w_{k}$ be the weight for the unit $k$ w.r.t. anomaly scoring, then based on \cite{zhou2016cam} we can obtain $\phi(\mathbf{x})=\sum_{i,j}\mathbf{M}(i,j)$, where $\mathbf{M}(i,j)=\sum_{k}w_{k}p_{k}(i,j)$ is the class activation map.
The frame-level saliency map can then be obtained by upsampling the class activation map to the size of the input frame $\mathbf{x}$. The CAM-based saliency maps corresponding to some exemplar anomalies are shown in Figure \ref{fig:saliency}. We can see that the regions corresponding to the anomalous events of the frames are well highlighted with high activation values in all four different scenes. Although our method may be also distracted by normal patches in some cases, such as the upper patch in Figure \ref{fig:saliency}(b), it is very impressive for an unsupervised anomaly detection method to achieve such effective anomalous region localization.

\subsection{Ablation Study}

\begin{figure}[h!]
  \centering
    \includegraphics[width=0.544\textwidth]{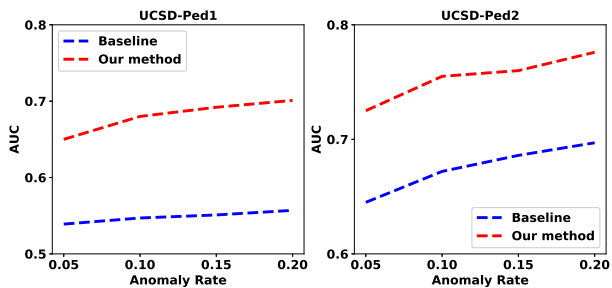}
  \caption{AUC performance w.r.t.\  different anomaly rates.}
  \label{fig:initial}
\end{figure}

\textbf{Initial Anomaly Detection}. The stability of labeling pseudo anomalies and normal data using the two fixed cut-off thresholds is examined by looking into the AUC performance on datasets with different anomaly rates, including 5\%, 10\%, 15\% and 20\%. The results on Ped1 and Ped2 are shown in Figure \ref{fig:initial}, with Sp + iForest used as baseline. This experiment is not applicable on the other datasets as their anomaly rates are too small. The results show that, despite the anomaly rate varies significantly across the examined cases, our method with the default cut-offs can consistently achieve substantial AUC improvement over the baseline that we use to generate initial pseudo labels. This frees us from tuning the cutoffs on different scenarios. Note that the increasing performance with increasing anomaly rates is mainly due to the fact that it becomes easier to obtain better AUC performance when the anomaly rate is larger.

\textbf{Network Architecture}. We replace ResNet-50 with VGG \cite{simonyan2015vgg} and 3DConv \cite{tran20153dconvolution} to examine the use of different architectures. The results are shown in Table \ref{tab:architecture}. It is clear that our method can work very well using different popular backbones with either shallower or higher-dimensional convolutional architectures. This indicates that our performance is not dependent on specific backbones.

\begin{table}[htbp]
\centering
\caption{ AUC performance of using different architectures.}
\scalebox{0.80}{
\begin{tabular}{cccccc}
\hline
\textbf{Backbone}  & \textbf{Ped1} & \textbf{Ped2} & \textbf{Entrance} & \textbf{Exit} & \textbf{UMN}\\
\hline
\textbf{VGG} & 70.4\% &  80.0\% &  86.5\% & 90.3\% & 97.4\%\\
\textbf{3DConv} &  70.1\% & 82.6\%  & 87.3\%  & 93.6\% & 98.1\%\\
\textbf{ResNet-50}  & 71.7\% &  83.2\% &  88.1\% & 92.7\% & 97.4\%\\
\hline
\end{tabular}
}
\label{tab:architecture}
\end{table}

\begin{figure}[h!]
  \centering
    \includegraphics[width=0.46\textwidth]{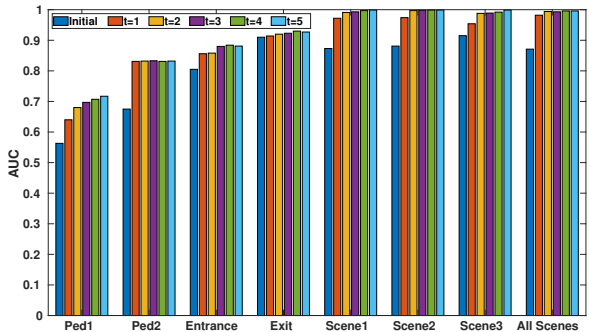}
  \caption{AUC performance of our method at each iteration. `$t=i$' indicates our model with $i$ iterations. }
  \label{fig:iteration}
\end{figure}

\textbf{Self-training}. To examine the self-training module, Figure \ref{fig:iteration} shows the AUC results of our method at each iteration during self-training. Our performance gets larger improvement with increasing iterations in the first few iterations on most datasets and then becomes stable at the 4th or 5th iteration. This shows that the self-training approach can iteratively improve the performance of our method. However, it should be noted that the performance of our method is bounded at some points when no extra information is provided, so we stop the iterative learning after a few iterations. We found empirically that five iterations are often sufficient to reach the possibly best performance on different datasets.

\textbf{End-to-end Anomaly Score Learner}. The importance of the end-to-end anomaly score learning is manifested by comparing the results of our method to Del Giorno et al. \cite{del2016discriminative} \#2 in Table \ref{tab:auconthreedata}.  The anomaly scoring of our method and Del Giorno et al. \cite{del2016discriminative} \#2 take exactly the same feature inputs, but our method achieves consistently large improvement over Del Giorno et al. \cite{del2016discriminative} \#2 on all the datasets. This is because the input features are optimized as an integrated part in our end-to-end score learning, resulting in optimal anomaly scores; whereas the two-step methods like Del Giorno et al. \cite{del2016discriminative} \#2 on one hand rely on the quality of input feature representations while on the other hand cannot unify the feature extractor/learner and anomaly scoring, leading to much less effective performance.

\section{Conclusions}

We have shown that framing video anomaly detection as a self-training deep ordinal regression task overcomes some of the key limitations of existing approaches to this important problem.  We additionally devised an end-to-end trained approach that outperforms the current state-of-the-art by a significant margin.
Two key insights gained are that (1) the end-to-end learning enables better optimized anomaly scores than the two-step approach and (2) the self-training ordinal regression approach can be leveraged by our end-to-end anomaly score learner to iteratively enhance detection performance.
Furthermore, our method offers some crucial capabilities, including human-in-the-loop anomaly detection and accurate anomaly localization. We are working on incorporating other features such as motion features into our model to identify other types of anomalies.

\textbf{Acknowledgments}
XB was in part supported by the NSFC project \#61772057 and BNSF project \#4202039.

{\small
\bibliographystyle{ieee_fullname}
\bibliography{reference}

}

\end{document}